\documentclass[letterpaper]{article} 
\usepackage{aaai25}  
\usepackage{times}  
\usepackage{helvet}  
\usepackage{courier}  
\usepackage[hyphens]{url}  
\usepackage{graphicx} 
\usepackage{natbib}  
\usepackage{caption} 
\usepackage{algorithm}
\usepackage{algorithmic}
\usepackage{newfloat}
\usepackage{listings}
\usepackage{amsmath}
\usepackage{amssymb}
\usepackage{bbding}
\usepackage{multirow}
\usepackage{newfloat}
\usepackage{listings}
\DeclareCaptionStyle{ruled}{labelfont=normalfont,labelsep=colon,strut=off} 
\floatstyle{ruled}
\newfloat{listing}{tb}{lst}{}
\floatname{listing}{Listing}

\title{RETQA: A Large-Scale Open-Domain Tabular Question Answering Dataset for Real Estate Sector}

\author{
    Zhensheng Wang\textsuperscript{\rm 1},Wenmian Yang\textsuperscript{\rm 2}\thanks{Corresponding authors.},Kun Zhou\textsuperscript{\rm 1},Yiquan Zhang\textsuperscript{\rm 3},Weijia Jia\textsuperscript{\rm 2,4*}\\
}
\affiliations{
    \textsuperscript{\rm 1}School of Artificial  Intelligence, Beijing Normal University, Beijing 100875, PR China\\

    \textsuperscript{\rm 2}Institute of Artificial Intelligence and Future Networks, Beijing Normal University, Zhuhai 519087, PR China

    \textsuperscript{\rm 3}Elmleaf Ltd.,  Shanghai 200082, PR China

    \textsuperscript{\rm 4}BNU-UIC Institute of Artificial Intelligence and Future Networks, Beijing Normal University (Zhuhai), Guangdong Key Lab of AI and Multi-Modal Data Processing, BNU-HKBU United International College, Zhuhai, Guang Dong, PR China

\{jensenwang,zhoukun\}@mail.bnu.edu.cn,
\{wenmianyang,jiawj\}@bnu.edu.cn,
zhangyq987@hotmail.com
}
\usepackage{bibentry}
\begin{document}
\maketitle

\begin{abstract}

The real estate market relies heavily on structured data, such as property details, market trends, and price fluctuations. However, the lack of specialized Tabular Question Answering datasets in this domain limits the development of automated question-answering systems. To fill this gap, we introduce RETQA, the first large-scale open-domain Chinese Tabular Question Answering dataset for Real Estate. RETQA comprises 4,932 tables and 20,762 question-answer pairs across 16 sub-fields within three major domains: property information, real estate company finance information and land auction information. Compared with existing tabular question answering datasets, RETQA poses greater challenges due to three key factors: long-table structures, open-domain retrieval, and multi-domain queries. To tackle these challenges, we propose the SLUTQA framework, which integrates large language models with spoken language understanding tasks to enhance retrieval and answering accuracy. Extensive experiments demonstrate that SLUTQA significantly improves the performance of large language models on RETQA by in-context learning. RETQA and SLUTQA provide essential resources for advancing tabular question answering research in the real estate domain, addressing critical challenges in open-domain and long-table question-answering.

\end{abstract}

\begin{links}
\link{Code}{https://github.com/jensenw1/RETQA}
\end{links}

\section{Introduction}

\begin{figure*}[htbp]
 \centering
 \includegraphics[width=5.5in]{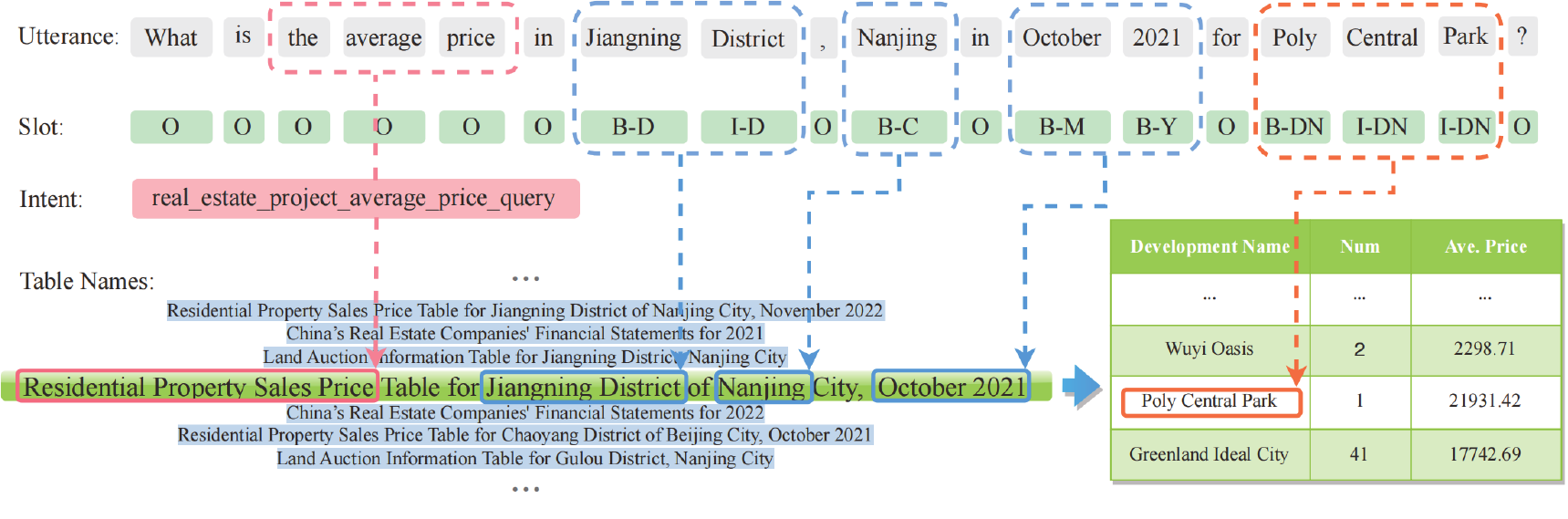}
 \centering
 \caption{Example of SLU labels and the relation to open-domain TQA, where ``D'' represents district, ``C'' represents city, ``M'' represents month, ``Y'' represents year, and ``DN'' represents development name.}
 \label{fig:Search table name by SLU}
\end{figure*}

With rapid advancements in artificial intelligence and natural language processing, Tabular Question Answering (TQA) has garnered attention for its ability to extract accurate answers from structured data across domains like finance and healthcare \cite{zhu2024zero}. Despite progress, specialized datasets for the real estate domain are scarce, limiting research and practical applications. The real estate market relies on structured data, such as property details and market trends, which are crucial for stakeholders like homebuyers and investors. For instance, homebuyers can make informed decisions by querying historical transaction prices and property details, while investors can assess the viability of investments by querying land information, asset status, and the developer’s transaction history. However, the lack of tailored TQA datasets in this field hampers the development of automated QA systems.

To address this gap, we introduce RETQA, the first large-scale open-domain Chinese TQA dataset specifically designed for the Real Estate sector. RETQA is built from publicly available real estate data and comprises 4,932 tables and 20,762 QA pairs spanning 16 sub-fields within three major domains: property information, real estate company finance, and land auction information. Given the complex and data-intensive nature of the real estate market, RETQA is crafted to generate queries targeting longer tables, with an average of 252.9 rows per query-related table. 
Moreover, queries in RETQA may related to more than one table. The above design captures the real estate domain's complexity, making RETQA a challenging TQA dataset.

The open-domain nature of RETQA requires models to retrieve relevant tables from the entire dataset, rather than being directly provided, effectively mirroring real-world scenarios. However, this process makes open-domain TQA more challenging than closed-domain, as retrieval accuracy directly affects overall TQA performance.
To facilitate retrieval, RETQA assigns a summary caption to each table. Furthermore, RETQA integrates the labels of Spoken Language Understanding (SLU) for each query. These SLU labels, including intent and slot labels, commonly employed in task-oriented dialogue systems \cite{qin2021survey,cheng2023scope,zhu2024zero,qin2024croprompt}, are instrumental in discerning user query intent and extracting pertinent details. As depicted in Figure \ref{fig:Search table name by SLU}, SLU labels enable more accurate parsing of user intent and relevant information, thereby enabling more precise retrieval and accurate answering. For each query, RETQA provides intent labels, slot labels, table captions, and answers in Markdown, SQL-style, and natural language formats. To the best of our knowledge, RETQA is the first TQA dataset to integrate SLU labels.

RETQA is created through four key steps: First, we compile a dataset of 4,932 real estate-related tables from eight major Chinese cities—Beijing, Shanghai, Guangdong, Shenzhen, Suzhou, Hangzhou, Nanjing, and Wuhan—spanning the years 2019 to 2022. These tables include property information, real estate company finances, and land auction data. The publicly available data is meticulously cleaned and organized, and each table is assigned a summary caption to aid retrieval. Second, we develop 90 real estate-specific question templates, addressing factual, inferential, and comparative queries. These templates are populated with real data to generate accurate QA pairs, with answers provided in Markdown, SQL-style, and natural language formats. Third, we annotate intent and slot labels by reverse engineering,  classifying queries into 16 intent categories and identifying six slot categories based on entity types found in the query templates and table headers. Finally, we use in-context learning to rewrite the template-generated questions, enhancing their naturalness and diversity. After filtering, RETQA retains 20,762 QA pairs.

In summary, RETQA accurately mirrors the real-world challenges users encounter when querying real estate-related data, including open-domain queries, long-table queries, and multi-domain and multi-table queries, making RETQA a particularly demanding task. To address these challenges, we propose SLUTQA as a benchmark for RETQA, which improves the performance of large language models (LLMs) on TQA tasks by leveraging SLU labels through in-context learning (ICL).

The SLUTQA framework consists of three key modules: the SLU module, the SLU label-based Retrieval (SR) module, and the SLU label-based Filtering-Answer (SFA) module. For each query, the SLU module generates intent and slot labels at first. Depending on the availability of SLU labels, we use two approaches: (1) fine-tuning a BERT \cite{devlin2019bert} model when labels are sufficient, and (2) employing in-context learning with large language models (LLMs) when labels are limited (few-shot). The SR module then uses those SLU labels to create a query summary via in-context learning, improving retrieval accuracy by replacing the original query when searching for the relevant table caption by BM25 \cite{robertson1994some}. Finally, the SFA module generates accurate SQL statements or refined Markdown answers based on the predicted SLU labels and retrieved tables. This framework leverages the strengths of large language models while integrating SLU task labels, enhancing the performance of open-domain long-table question answering tasks. To the best of our knowledge, this is the first work to combine SLU tasks with TQA tasks.

Extensive experiments on RETQA demonstrate that SLUTQA significantly enhances LLM performance without requiring fine-tuning. Our dataset and framework offer valuable resources for advancing TQA research in the real estate domain, promoting further development in this field. The code and data will be released after the blind review. The main contributions of this paper are as follows:

\begin{itemize}
\item We introduce the first large-scale open-domain Chinese TQA dataset in the real estate domain, enriched with SLU labels. This dataset is challenging due to its long-table, multi-table, and open-domain characteristics.
\item We propose the SLUTQA framework, which integrates LLMs with SLU tasks, significantly improving TQA accuracy on the RETQA dataset. To the best of our knowledge, this is the first work to combine SLU tasks with table QA tasks.
\item Extensive experimental results show that SLUTQA improves the performance of existing LLMs on RETQA without the need for fine-tuning.
\end{itemize}

\begin{table*}[htbp]
\centering
\begin{tabular}{cccccccc}
\hline
Dataset  & Open Domain  & \# of QA pairs & \# Tables & Answer format  & Multi-table  & SLU          & Long Table \\ \hline
WikiSQL           & \XSolidBrush & 80654          & 24241     & SQL      & \XSolidBrush & \XSolidBrush & \XSolidBrush     \\
WikiTableQuestion & \XSolidBrush & 22033          & 2108      & Text     & \XSolidBrush & \XSolidBrush & \XSolidBrush     \\
Spider            & \XSolidBrush & 10181          & 1020      & SQL      & \Checkmark   & \XSolidBrush & \Checkmark         \\
Open-WikiTable    & \Checkmark   & 67023          & 24680     & Text,SQL & \XSolidBrush & \XSolidBrush & \XSolidBrush     \\
NQ-TABLES         & \Checkmark   & 11628          & 169898    & Text     & \XSolidBrush & \XSolidBrush & \textbf{--}         \\ \hline
RETQA (Ours)              & \Checkmark   & 20762          & 4932      & Text,SQL & \Checkmark   & \Checkmark   & \Checkmark     \\ \hline
\end{tabular}
\caption{Dataset Comparison, where ``\textbf{--}'' indicates data currently inaccessible. We define a dataset as a long table dataset if the average number of rows in the query-related tables exceeds 100.}
\label{Compare to other}
\end{table*}

\section{Related Works}
Our work is related to two areas, i.e., TQA datasets, and TQA methods.

\subsection{TQA Datasets}
Early research, such as WikiTableQuestions \cite{pasupat2015compositional} and WikiSQL \cite{zhong2017seq2sql}, collected tabular data from web sources like Wikipedia. However, WikiTableQuestions only provides text answers, and WikiSQL's question descriptions are too vague to reliably locate relevant tables. The Spider dataset \cite{yu2018spider}, designed for complex, cross-domain Text-to-SQL tasks, is not suited for open-domain scenarios and only offers SQL-format answers. NQ-TABLES \cite{herzig2021open}, the first open-domain table QA dataset, uses a reader model to extract answers from $K$ candidate tables by selecting a single cell, limiting the need for complex reasoning \cite{kweon2023open}. Open-WikiTable \cite{kweon2023open}, based on WikiSQL and WikiTableQuestions, offers both SQL and text answers but typically features short tables. To the best of our knowledge, we are the first TQA dataset providing the SLU labels. The comparison of RETQA to other related datasets is shown in Table \ref{Compare to other}.

\subsection{TQA Methods}
Before the emergence of large language models (LLMs), researchers explored methods to combine tabular data with neural networks for natural language processing and data management tasks \cite{badaro2023transformers,fang2024large,shwartz2022tabular}. LLMs gained attention for their strong cross-task generalization, adapting to new tasks with minimal examples. \cite{chen2023large} first demonstrated LLMs' ability to reason over tables through in-context learning. TAP4LLM \cite{sui2023tap4llm} addresses noisy table data by enhancing sub-tables. OPENTAB retrieves tables, generates SQL as intermediate steps, and relies on a reader for final answers, though it may produce incorrect SQL \cite{kong2024opentab}. StructGPT \cite{jiang2023structgpt} uses interfaces to handle large structured data efficiently. MultiTabQA \cite{pal2023multitabqa} introduces an LLM-based framework for answering questions across multiple tables, but it is limited to queries in closed domains. In comparison, our novel integration of LLMs with SLU tasks significantly enhances open-domain TQA accuracy.

\section{Dataset Construction and Analysis}
This section elaborates on how RETQA is created, including details on table collection, QA pair generation, intent and slot annotation, and query rewriting and quality control.
\subsection{Table Collection}
Our table data comes from publicly available real-world sources, including property data, real estate company financial data, and land auction data. The specific data sources and descriptions are as follows:

\noindent\textbf{Property Data:} We collected commercial housing transaction data from eight major Chinese cities (Beijing, Shanghai, Guangzhou, Shenzhen, Suzhou, Hangzhou, Nanjing, Wuhan) for 2019-2022, sourced from http://www.fangdi.com.cn/. The tables, keyed by development name, include district, average transaction prices, year and month of transaction dates, number of transactions, developers, etc. This resulted in 4,825 standardized tables, captioned with the administrative region and year-month (e.g., ``Residential Property Sales Price Table for Jiangning District of Nanjing City, May 2022'').

\noindent\textbf{Real Estate Company Finance Data:} Financial disclosures from Chinese real estate companies for 2019-2022 are collected from https://aur.elmleaf.com.cn. Each table, keyed by company name, includes total operating revenue, operating profit, total operating costs, total assets, total liabilities, state-owned status, credit bond status, and risk level. The data is organized by year, resulting in four tables captioned with the year (e.g., ``China's Real Estate Companies' Financial Statements for 2020'').

\noindent\textbf{Land Auction Data:} Land auction information from 2016-2022 for the same eight cities is collected from city-specific public websites: Beijing\footnote{https://zjw.beijing.gov.cn/}, Shanghai\footnote{https://biz.ghzyj.sh.gov.cn/shtdsc/wz}, Guangzhou\footnote{https://zfcj.gz.gov.cn/zfcj/fyxx/fdcxmxx}, Shenzhen\footnote{http://zjj.sz.gov.cn:8004/}, Suzhou\footnote{http://112.80.51.227:6600/index}, Hangzhou\footnote{https://www.zjzrzyjy.com/portalBrowse/home}, Nanjing\footnote{http://www.landnj.cn}, and Wuhan\footnote{http://www.whtdsc.com}. Tables, keyed by land parcel name, include details like development name, transaction date, affiliated group, total transaction price, floor area ratio, building density (\%), green coverage ratio (\%), etc. This yielded 103 tables, captioned by district(e.g., ``Land Auction Information Table for Xuhui District, Shanghai'').

In total, our dataset integrates 4,932 tables across these three domains, with the longest table containing 465 rows. Each table is annotated with a standardized caption for retrieval. For more details about the table, see Appendix C.

\subsection{QA Pair Generation}

We utilize a template-based approach to generate QA pairs automatically. Specifically, we define 90 query templates covering various types of inquiries, including factual, inferential, and comparative questions, derived from extracted tables and create corresponding SQL templates for each query. Among these, 23 templates are designed to generate queries that require multi-table support. Given the complex and data-intensive nature of the real estate market, we focus on generating queries that target longer tables to ensure that the dataset accurately reflects this complexity. Consequently, the average number of rows per query-related table is 252.9. Additional details on the templates are provided in Appendix A.

Using these templates, we generated 300 natural language queries and corresponding SQL statements per template, resulting in a total of 27,000 pairs. We then filtered out QA pairs with duplicate queries, non-executable SQL statements, and SQL statements that return empty results, leaving us with 20,471 QA pairs. For each remaining pair, we also provide a matching Markdown-formatted input table (with multiple tables concatenated into a single row) and the corresponding output answer table. Since the real estate data is collected from Chinese public websites, all queries are generated in Chinese. We plan to publish an English version soon.

\subsubsection{Intent and Slot Annotation}

To facilitate researchers effectively extracting key information from queries and achieve accurate retrieval and answers, RETQA includes additional Spoken Language Understanding (SLU) labels. SLU consists of two types of labels: intent and slot labels, commonly used in task-oriented dialogue systems. Intent labels identify the user’s query intent, while slot labels extract key information pertinent to that intent.

We annotate the intent and slot labels through reverse engineering. Specifically, for intent labels, we categorized the queries into 16 types based on the 90 templates mentioned earlier. Some queries in the templates may involve multiple intents. For example, in the query, ``Which of the top 5 best-selling residential compounds in Bao'an District, Shenzhen, has the lowest housing price'', both average price and sales volume information are required. In this case, the query would be labeled with two intents: ``real estate project sales volume query'' and ``real estate project average transaction price query.'' For more details and statistics about the intent labels, please see Appendix C.

For slot labels, we categorize the entities into six types: ``city'', ``district'', ``development name'', ``company name'', ``year'', and ``month',' based on the entity types in the query templates and corresponding table headers. Following previous work on SLU tasks \cite{qin2022multi}, we adopt the Inside–Outside–Beginning (IOB) tagging format \cite{ramshaw1999text}. In this format, the B- prefix indicates the beginning of a slot chunk, the I- prefix indicates that the tag is inside a slot chunk, and the O tag indicates that a token does not belong to any slot chunk. An example of SLU labels is shown in Figure \ref{fig:Search table name by SLU}.

\subsubsection{LLM-based Query Rewriting and Quality Control}

Despite our extensive template library, the sentences generated from these templates can be grammatically monotonous, differing significantly from real human queries. This disparity may cause models trained on template-based data to struggle with actual user queries. To address this, we use large language models (LLMs) to rewrite the queries generated by templates, creating more diverse and human-like expressions while retaining their original meaning. Techniques such as synonym replacement and sentence inversion are employed in this process.

We used Qwen2 72B \cite{yang2024qwen2} for rewriting and DeepSeek-V1 (Dense-67B) \cite{deepseek-llm} to evaluate the results. The evaluation scores questions on a scale from 0 (``completely like template generation'') to 5 (``like human writing''). After rewriting, the average score increased from 2.56 to 2.95, indicating a significant improvement in naturalness and diversity. The distribution of scores before and after rewriting is shown in Appendix B.

After rewriting, we manually reviewed and removed 291 incorrect QA pairs, leaving 20,762 pairs in the RETQA dataset, and split them into training, testing, and validation sets at a ratio of 0.8:0.1:0.1. To ensure that each template was represented in each set, we perform sampling based on the templates. Finally, for each query, RETQA provides the corresponding intent labels, slot labels, and the caption of the relevant table(s), as well as answers in three formats: Markdown, SQL-style (with both the SQL statements and their resulting answers), and natural language. Detailed statistics of RETQA are shown in Appendix C.

\section{Method}
In this section, we introduce the SLUTQA framework, which consists of three key modules:  the SLU module, the SLU label-based Retrieval (SR) module, and the SLU label-based Filtering-Answer (SFA) module. SLUTQA aims to enhance the performance of LLMs on TQA tasks through SLU labels-based in-context learning (ICL). The general framework of SLUTQA is shown in Figure \ref{fig:Pipeline}.

\begin{figure*}[htbp]
 \centering
 \includegraphics[width=0.95\textwidth]{figures/pipeline_cmyk.pdf}
 \centering
 \caption{General framework of SLUTQA.}
 \label{fig:Pipeline}
\end{figure*}

\subsection{SLU Module}

SLU involves two label types, i.e., intent and slot, commonly used in task-oriented dialogue systems. Intent labels identify the user's query intent, while slot labels extract relevant key information. In this paper, we design a SLU module to predict the SLU labels of each query, and leverage these labels to enhance information extraction and ensure accurate retrieval and answers within the SLUQTA framework.

Based on the availability of labeled SLU data, we design two approaches to implement the SLU module: (1) fine-tuning a BERT model \cite{devlin2019bert} when sufficient labeled data are available, and (2) utilizing ICL with LLMs in scenarios where labeled data are limited (few-shot).

In the first scenario, given the current query $\textbf{X} = \{x_1, ..., x_n \}$ as input, we utilize a pre-trained BERT \footnote{https://huggingface.co/google-bert/bert-base-uncased} as the encoder. The model returns the hidden states $H = \{h_{cls}, h_1, ..., h_n \in \mathbb{R}^{d_{model}} \}$, where CLS is a special token representing the entire input sequence within BERT, and $ d_{model}$ is the output dimension of BERT.

Then, we predict the intent labels, number of intents, and slot labels by:
\begin{equation}
\begin{aligned}
\text{y}^I &= \mathrm{Sigmoid}(\mathbf{W}^I \cdot \mathbf{h}_{cls} + \mathbf{b}^I), \\[6pt]
\text{y}^N &= \mathrm{Softmax}(\mathbf{W}^N \cdot \mathbf{h}_{cls} + \mathbf{b}^N), \\[6pt]
\text{y}_j^S &= \mathrm{Softmax}(\mathbf{W}^S \cdot \mathbf{h}_j + \mathbf{b}^S),
\end{aligned}
\end{equation}
where $\mathbf{y}^I\in\mathbb{R}^{d_i} $, $ \mathbf{y}^N\in \mathbb{R}^{d_n} $, and $ \mathbf{y}^S={\mathbf{y}_1^S,...,\mathbf{y}_n^S\in\mathbb{R}^{d_s}} $ represent the predicted results of intents, number of intent, and slots, respectively, $\mathbf{W}^I\in\mathbb{R}^{d_i\times d_{model}} $, $ \mathbf{W}^N\in\mathbb{R}^{d_n\times d_{model}} $, and $ \mathbf{W}^S\in\mathbb{R}^{d_s\times d_{model}} $ are fully connected matrices, $ \mathbf{b}^I\in\mathbb{R}^{d_i} $,$ \mathbf{b}^N\in\mathbb{R}^{d_n} $ and $ \mathbf{b}^S\in\mathbb{R}^{d_s} $ are bias vectors. In this paper, $d_i=16$, $d_n=2$, and $d_s=13$ (6 kinds of slots with B-tags, I-tags, and the O-tag), representing the categories of intent labels, number of intents, and categories of slot labels, respectively.

During fine-tuning, cross-entropy loss is employed for both intent number prediction and slot filling, while binary cross-entropy is used for intent detection. During inference, the top 1 or 2 predicted intents are selected based on the predicted number of intents.

In the second scenario, where labeled SLU data are limited (few-shot) and insufficient for training or fine-tuning a model, we predict SLU labels using ICL in LLMs. Specifically, we randomly select 22 examples from the training set to serve as prompt examples for the LLMs. As illustrated in Figure \ref{fig:Pipeline} (top left, SLU Examples), the LLM leverages these few-shot examples as prompts to jointly predict intent and slot labels for each input query.

\subsection{SR Module}

Although our dataset provides corresponding tables (i.e., gold tables) for each query, in real-world open-domain scenarios, these tables must be retrieved based on the query, and the retrieval accuracy directly affects overall TQA performance. Previous works \cite{kong2024opentab} often use BM25 \cite{robertson2009probabilistic} for table retrieval due to its scalability and competitive performance. However, in the real estate data scenario, user queries can be diverse and non-standard, while table captions are often similar (e.g., differing only by years or months). This means that relying solely on BM25 for table retrieval may not capture the critical information needed, leading to lower accuracy in our scenario.

To address this issue, we propose the SR module, which utilizes ICL with SLU labels to enable the large LLMs to generate a query summary in the form of a table caption. For each query, five examples from the training set that share the same intent (as predicted by the SLU module) are selected and used as prompts for the LLM. As illustrated in Figure \ref{fig:Pipeline} (bottom left, SR Examples), each example comprises four components: the query, intent(s), slots, and query summary, with the query summary corresponding to the table caption(s) in the example, which may include multiple tables. This approach enables the LLM to learn to generate a summary in the form of a table caption relevant to the query. The intent and slot information allow the LLM to more accurately capture the key details in the query, resulting in a more precise summary. Finally, the SR-generated summary, rather than the original query, is used to retrieve the table caption via BM25. It is important to note that the generated summary may consist of multiple entries, with each summary being used to retrieve one table using BM25, selecting the top result for each.

\begin{table*}[htbp]\small
\begin{tabular}{clcccccccccc}
\hline
\multirow{2}{*}{Model}     & \multicolumn{1}{c}{\multirow{2}{*}{Method}} & \multirow{2}{*}{Table EM(\%)} & \multicolumn{3}{c}{Row EM(\%)}                   & \multicolumn{3}{c}{Column EM(\%)}                & \multicolumn{3}{c}{Cell EM(\%)}                  \\ \cline{4-12} 
                           & \multicolumn{1}{c}{}                        &                               & P              & R              & F1             & P              & R              & F1             & P              & R              & F1             \\ \hline
\multirow{3}{*}{Qwen2 7b}  & Vanilla                                     & 15.05                         & 24.43          & \textbf{31.71} & 27.60          & \textbf{25.10} & 22.11          & 23.51          & 36.19          & \textbf{43.35} & 39.45          \\
                           & SLUTQA (ICL)                                    & 16.65                         & 26.73          & 25.34          & 26.02          & 23.36          & 26.06          & 24.64          & 40.67          & 40.45          & 40.56          \\
                           & SLUTQA (FT)                                  & \textbf{18.94}                & \textbf{32.36} & 29.61          & \textbf{30.92} & 23.14          & \textbf{28.90} & \textbf{25.70} & \textbf{45.07} & 43.24          & \textbf{44.14} \\ \hline
\multirow{3}{*}{GLM4   9b} & Vanilla                                     & 4.73                          & 26.55          & \textbf{45.71} & 33.59          & \textbf{31.17} & 30.36          & 30.76          & 35.77          & \textbf{59.42} & 44.66          \\
                           & SLUTQA (ICL)                                    & \textbf{9.15}                 & 34.88          & 40.42          & 37.44          & 28.19          & \textbf{34.25} & 30.92          & 44.66          & 55.80          & 49.61          \\
                           & SLUTQA (FT)                                    & 9.08                          & \textbf{37.62} & 43.44          & \textbf{40.32} & 28.73          & 33.93          & \textbf{31.11} & \textbf{47.75} & 58.01          & \textbf{52.38} \\ \hline
\multirow{3}{*}{Qwen2 72b} & Vanilla                                     & 11.14                         & 36.27          & 54.54          & 43.57          & 32.16          & 32.93          & 32.54          & 44.64          & 58.46          & 50.62          \\
                           & SLUTQA (ICL)                                     & 14.81                         & 49.79          & 53.74          & 51.69          & \textbf{44.01} & \textbf{42.75} & \textbf{43.37} & 64.51          & 67.68          & 66.06          \\
                           &  SLUTQA (FT)                                & \textbf{15.02}                & \textbf{51.26} & \textbf{55.61} & \textbf{53.34} & 43.39          & 41.90          & 42.63          & \textbf{65.01} & \textbf{68.29} & \textbf{66.61} \\ \hline
\end{tabular}
\caption{Overall performance of markdown format answer.}\label{Markdown Answer Result}
\end{table*}

\subsection{SFA Module}
After retrieving the corresponding tables from the SR module, we designed the SFA module to generate final answers in two formats: SQL-style and markdown.

In markdown formats, existing methods typically flatten entire tables into a single row for processing by large language models (LLMs). However, RETQA includes many lengthy tables, and current LLMs, with their limited input lengths (usually 32k tokens), struggle to effectively handle queries involving these extensive tables. To address this issue, the SFA module for markdown formats is designed to filter out rows and columns that are irrelevant to the query. This is achieved with the help of SLU labels, which contain key information that guides the LLM to focus on the relevant portions of the table. Specifically, the SFA module operates in two steps. First, the original markdown table is simplified by removing irrelevant rows and columns based on the identified slots and intent. Second, the simplified table is used to generate the final output. This process is facilitated through ICL. For each query, 2-5 examples (according to the total length) from the training set, all sharing the same intent, are selected. As illustrated in Figure \ref{fig:Pipeline} (bottom right, SFA Markdown Examples), the first stage of the process includes examples consisting of five components: the query, intent(s), slots, original table, and simplified table. In the second stage, the examples include the query, intent(s), slots, simplified table, and the final answer.

For SQL-style answers, the primary challenge is accurately extracting key information from queries to generate the corresponding SQL statements. To address this, the SLA module for SQL formats leverages SLU labels and ICL to generate SQL statements. As with the markdown process, we select five examples from the training set for each query, all sharing the same intent. As illustrated in Figure \ref{fig:Pipeline} (top right, SFA SQL Examples), these examples consist of the query, intent(s), slots, retrieved table caption(s), and SQL statements. The final output is produced by executing the generated SQL statement.

\begin{table}[t]
\centering
\begin{tabular}{clcc}
\hline
Model                       & \multicolumn{1}{c}{Method} & ECR(\%)        & pass@1 \\ \hline
                            & Vanilla                    & 60.11          & 39.55                               \\
                            & SLUTQA (ICL)                 & 85.46          & 64.46                               \\
\multirow{-3}{*}{Qwen2 7b}  & SLUTQA (FT)                & \textbf{87.28} & \textbf{70.16}                      \\ \hline
                            & Vanilla                    & 68.49          & 51.40                               \\
                            & SLUTQA (ICL)                  & 92.06          & 75.65                               \\
\multirow{-3}{*}{Qwen2 72b} & SLUTQA (FT)                 & \textbf{97.23} & \textbf{82.71}                      \\ \hline
                            & Vanilla                    & 56.62          & 34.52                               \\
                            & SLUTQA (ICL)                  & 84.97          & 60.99                               \\
\multirow{-3}{*}{GLM4   9b} & SLUTQA (FT)                & \textbf{89.90} & \textbf{66.81}                      \\ \hline
\end{tabular}
\caption{Overall performance of SQL format answer.}\label{Final result of SQL}
\end{table}

\section{Experiments}
In this section, we first introduce the experiment setup. Then, we show the experiment results and conduct ablation studies.

\subsection{Experimental Settings and Baselines}

\textbf{Evaluation Metrics:} 
In this paper, we provide answers to each query in three formats: Markdown, SQL-style, and natural language. However, since natural language output is subjective, we focus on evaluating the two objective formats, Markdown and SQL-style.

For SQL-style answers, we assess performance using Executable Code Ratio (ECR) and Pass Rate (pass@1), as outlined in \cite{he2024text2analysis}. For Markdown answers, we evaluate using Table Exact Match accuracy (Table EM) and Precision (P), Recall (R), and F1 score for exact matches of rows, columns, and cells, following the metrics described in \cite{pal2023multitabqa}. Additionally, for table retrieval performance, we also employ Precision, Recall, and F1 score as evaluation metrics.

\noindent\textbf{Baselines:} To prove that our designed SLUTQA could enhance the performance of existing LLMs on the RETQA dataset, we utilize our framework on three GPT family models, i.e., Qwen2-7b, Qwen2-72b \cite{yang2024qwen2}, and Glm4-9b \cite{glm2024chatglm}.

\subsection{Result and Analysis}

\textbf{Main Results:}
To evaluate the performance of SLUTQA on RETQA, we generated markdown and SQL-style answers using baseline LLMs through two approaches: the vanilla implementation and our SLUTQA framework. In the vanilla implementation, we generated a query summary for BM25 retrieval and then produced the final answers using ICL with five randomly selected examples. The key difference is that the vanilla approach does not include SLU labels and lacks a simplification step for markdown formatting.

The overall performance of the markdown and SQL-style answers is presented in Tables \ref{Markdown Answer Result} and \ref{Final result of SQL}, respectively. In these tables, “Vanilla” refers to the baseline implementation, “SLUTQA (ICL)” denotes the SLUTQA framework with the SR module implemented via in-context learning (ICL), and “SLUTQA (FT)” represents the SLUTQA framework with the SR module implemented by fine-tuning a BERT model. The results clearly demonstrate that our SLUTQA framework outperforms all baseline methods in both markdown and SQL formats, and further confirms that incorporating SLU labels can significantly enhance the performance of LLMs in open-domain TQA tasks.

Specifically, for markdown format answers, comparing SLUTQA (FT) with the vanilla implementation, Table EM accuracy improves by 3.89\%, 4.35\%, and 3.88\% for Qwen2 7b, GLM4 9b, and Qwen2 72b, respectively. Even when comparing SLUTQA (ICL), which does not fine-tune with a large number of SLU labels, with the vanilla implementation, Table EM accuracy still improves by 1.6\%, 4.42\%, and 3.67\% for Qwen2 7b, GLM4 9b, and Qwen2 72b, respectively. For SQL format answers, both SLUTQA (FT) and SLUTQA (ICL) outperform the vanilla implementation by more than 20\% on both ECR and pass@1 scores.

These results indicate that: (1) Current LLMs are capable of generating accurate SLU labels with just few-shot (22 in practice) examples. (2) SLU labels facilitate more precise parsing of user intent and relevant information, significantly enhancing the overall performance of open-domain TQA tasks. Detailed results of the SLU tasks are provided in Appendix D. This also demonstrates that incorporating traditional SLU labels into open-domain TQA tasks is a valid and effective approach, offering valuable insights for future TQA research.

\begin{table}[t]
\centering
\begin{tabular}{ccccc}
\hline
\multicolumn{2}{c}{Method}           & P              & R              & F1             \\ \hline
BM25                       & Top 1   & 76.86          & 56.02          & 64.80           \\ \hline
\multirow{3}{*}{Qwen2 7B}  & Vanilla & 93.87          & 89.09          & 91.42          \\
                           & SR(ICL) & 95.80          & 92.32          & 94.03          \\
                           & SR(FT)  & \textbf{97.53} & \textbf{95.05} & \textbf{96.27} \\ \hline
\multirow{3}{*}{Qwen2 72B} & Vanilla & 97.23          & 96.30          & 96.76          \\
                           & SR(ICL) & 97.66          & 97.35          & 97.50          \\
                           & SR(FT)  & \textbf{97.85} & \textbf{97.57} & \textbf{97.71} \\ \hline
\multirow{3}{*}{GLM4 9B}   & Vanilla & 93.73          & 92.02          & 92.87          \\
                           & SR(ICL) & 95.12          & 92.75          & 93.92          \\
                           & SR(FT)  & \textbf{97.67} & \textbf{94.29} & \textbf{95.95} \\ \hline
\end{tabular}
\caption{Table retrieval performance of different methods.}\label{Compare of Table Name predict}
\end{table}

\begin{table}[t]
\centering
\begin{tabular}{clccc}
\hline
\multirow{2}{*}{Model}     & \multicolumn{1}{c}{\multirow{2}{*}{Method}} & \multicolumn{3}{c}{Cell EM(\%)}                  \\ \cline{3-5} 
                           & \multicolumn{1}{c}{}                        & P              & R              & F1             \\ \hline
\multirow{3}{*}{Qwen2 7b}  & Vanilla                                     & 37.67          & 43.58          & 40.41          \\
                           & Simplified                                  & 34.07          & 37.88          & 35.87          \\
                           & SFA                                         & \textbf{39.62} & \textbf{44.02} & \textbf{41.71} \\ \hline
\multirow{3}{*}{GLM4 9b}   & Vanilla                                     & 39.85          & 61.96          & 48.51          \\
                           & Simplified                                  & 42.25          & 57.75          & 48.80          \\
                           & SFA                                         & \textbf{51.89} & \textbf{62.87} & \textbf{56.85} \\ \hline
\multirow{3}{*}{Qwen2 72b} & Vanilla                                     & 45.72          & 58.97          & 51.51          \\
                           & Simplified                                  & 62.35          & 69.20          & 65.59          \\
                           & SFA                                         & \textbf{66.36} & \textbf{70.27} & \textbf{68.26} \\ \hline
\end{tabular}
\caption{Ablation study of SFA module in markdown format.}\label{SLU filter ablation markdown experimence}
\end{table}

\begin{table}[htbp]
\centering
\begin{tabular}{clll}
\hline
Model                       & \multicolumn{1}{c}{Method} & ECR(\%)        & pass@1 \\ \hline
                            & Vanilla                   & 64.99          & 42.24                               \\
\multirow{-2}{*}{Qwen2 7b}  & SFA       & \textbf{90.17} & \textbf{74.58}                      \\ \hline
                            & Vanilla                   & 71.76          & 53.70                               \\
\multirow{-2}{*}{Qwen2 72b} & SFA      & \textbf{98.86} & \textbf{86.26}                      \\ \hline
                            & Vanilla                  & 60.01          & 35.25                               \\
\multirow{-2}{*}{GLM4 9b}   & SFA       & \textbf{90.36} & \textbf{69.65}                      \\ \hline
\end{tabular}
\caption{Ablation study of SFA module in SQL format.}\label{SLU filter ablation experimence}
\end{table}

\noindent\textbf{Ablation Study:} We conducted an ablation study to evaluate the effectiveness of the SR and SFA modules in SLUTQA.

First, we evaluated the SR module's impact on retrieval performance, as shown in Table \ref{Compare of Table Name predict}. We compared several methods: BM25, which retrieves the top 1 table based on the original query; the vanilla approach, which generates query summaries replacing the original query for BM25 retrieval; and our SR module. In SR (ICL) and SR (FT), summaries are generated using our SR module, with SLU labels predicted by ICL or a fine-tuned (FT) BERT model. The results show that BM25 performs significantly worse than the other methods, highlighting the effectiveness of LLM-generated table summaries for TQA tasks. Additionally, both SR (ICL) and SR (FT) outperform the vanilla approach across all LLMs, demonstrating that SLU labels enhance retrieval accuracy.

Next, we evaluated the SFA module. To isolate its impact, we used the gold table(s) and ground-truth SLU labels, ensuring the SFA module's performance is assessed independently of retrieval accuracy and SLU prediction.

The markdown results are shown in Table \ref{SLU filter ablation markdown experimence}. Due to the space limit, we only show the performance of exact matches of cells. Here ``Vanilla'' refers to the vanilla implementation described in the Main Results section, where the LLM generates the final answer from the query and gold markdown table(s) using in-context learning with 5 randomly selected examples. ``Simplified'' is an ICL-based method similar to the SFA module but without SLU labels in the prompts. ``SFA'' represents our SFA module. The results demonstrate that the SFA module outperforms both Vanilla and Simplified across all LLMs, indicating that SLU labels help simplify tables and reduce noise, addressing the challenges of long tables. Notably, Vanilla outperforms Simplified on Qwen2 7B, suggesting that without SLU labels, direct table simplification through ICL can inadvertently remove important information, reducing performance.

The SQL results are shown in Table \ref{SLU filter ablation experimence}. Again, ``Vanilla'' refers to the vanilla implementation in the Main Results section, while ``SFA'' denotes our SFA module. The results show that the SFA module significantly outperforms Vanilla across all LLMs, indicating that SLU labels help LLMs capture key information and generate more accurate SQL statements.

\section{Conclusion}
In this paper, we introduce RETQA, the first large-scale open-domain Chinese TQA dataset for the real estate sector, comprising 4,932 tables and 20,762 QA pairs. RETQA addresses the lack of specialized datasets by providing a challenging resource for TQA research, particularly in handling open-domain, long tables and multi-domain queries. To enhance performance, we developed the SLUTQA framework, which leverages LLMs and SLU labels for improved retrieval and answer generation. Extensive experiments show that SLUTQA significantly boosts LLM performance, offering valuable insights and resources for future TQA research.

\section{Acknowledgements}

This work was supported in part by the Chinese National Research Fund (NSFC) under Grant 62272050 and Grant 62302048; Zhuhai Science-Tech Innovation Bureau under Grant No. 2320004002772, and in part by the Interdisciplinary Intelligence SuperComputer Center of Beijing Normal University (Zhuhai). The acquisition of real estate data is supported by Elmleaf Limited (Shanghai).

\bibliography{myreference}

\begin{thebibliography}{26}
\providecommand{\natexlab}[1]{#1}

\bibitem[{Badaro, Saeed, and Papotti(2023)}]{badaro2023transformers}
Badaro, G.; Saeed, M.; and Papotti, P. 2023.
\newblock Transformers for Tabular Data Representation: A Survey of Models and
  Applications.
\newblock \emph{Transactions of the Association for Computational Linguistics},
  11: 227--249.

\bibitem[{Chen(2023)}]{chen2023large}
Chen, W. 2023.
\newblock Large Language Models are few (1)-shot Table Reasoners.
\newblock In \emph{Findings of the Association for Computational Linguistics:
  EACL 2023}, 1120--1130.

\bibitem[{Cheng, Yang, and Jia(2023)}]{cheng2023scope}
Cheng, L.; Yang, W.; and Jia, W. 2023.
\newblock A scope sensitive and result attentive model for multi-intent spoken
  language understanding.
\newblock In \emph{Proceedings of the AAAI Conference on Artificial
  Intelligence}, 12691--12699. {AAAI} Press.

\bibitem[{DeepSeek-AI(2024)}]{deepseek-llm}
DeepSeek-AI. 2024.
\newblock DeepSeek LLM: Scaling Open-Source Language Models with Longtermism.
\newblock \emph{arXiv preprint arXiv:2401.02954}.

\bibitem[{Devlin et~al.(2019)Devlin, Chang, Lee, and
  Toutanova}]{devlin2019bert}
Devlin, J.; Chang, M.-W.; Lee, K.; and Toutanova, K. 2019.
\newblock BERT: Pre-training of Deep Bidirectional Transformers for Language
  Understanding.
\newblock In \emph{Proceedings of the 2019 Conference of the North American
  Chapter of the Association for Computational Linguistics: Human Language
  Technologies, Volume 1 (Long and Short Papers)}, 4171--4186.

\bibitem[{Fang et~al.(2024)Fang, Xu, Tan, Zhang, Hu, Qi, Nickleach, Socolinsky,
  Sengamedu, Faloutsos et~al.}]{fang2024large}
Fang, X.; Xu, W.; Tan, F.~A.; Zhang, J.; Hu, Z.; Qi, Y.~J.; Nickleach, S.;
  Socolinsky, D.; Sengamedu, S.; Faloutsos, C.; et~al. 2024.
\newblock Large language models (LLMs) on tabular data: Prediction, generation,
  and understanding-a survey.
\newblock \emph{arXiv preprint arXiv:2402.17944}.

\bibitem[{GLM et~al.(2024)GLM, Zeng, Xu, Wang, Zhang, Yin, Rojas, Feng, Zhao,
  Lai et~al.}]{glm2024chatglm}
GLM, T.; Zeng, A.; Xu, B.; Wang, B.; Zhang, C.; Yin, D.; Rojas, D.; Feng, G.;
  Zhao, H.; Lai, H.; et~al. 2024.
\newblock ChatGLM: A Family of Large Language Models from GLM-130B to GLM-4 All
  Tools.
\newblock \emph{arXiv preprint arXiv:2406.12793}.

\bibitem[{He et~al.(2024)He, Zhou, Xu, Ma, Ding, Du, Gao, Jia, Chen, Han
  et~al.}]{he2024text2analysis}
He, X.; Zhou, M.; Xu, X.; Ma, X.; Ding, R.; Du, L.; Gao, Y.; Jia, R.; Chen, X.;
  Han, S.; et~al. 2024.
\newblock Text2analysis: A benchmark of table question answering with advanced
  data analysis and unclear queries.
\newblock In \emph{Proceedings of the AAAI Conference on Artificial
  Intelligence}, 18206--18215. {AAAI} Press.

\bibitem[{Herzig et~al.(2021)Herzig, Mueller, Krichene, and
  Eisenschlos}]{herzig2021open}
Herzig, J.; Mueller, T.; Krichene, S.; and Eisenschlos, J. 2021.
\newblock Open Domain Question Answering over Tables via Dense Retrieval.
\newblock In \emph{Proceedings of the 2021 Conference of the North American
  Chapter of the Association for Computational Linguistics: Human Language
  Technologies}, 512--519.

\bibitem[{Jiang et~al.(2023)Jiang, Zhou, Dong, Ye, Zhao, and
  Wen}]{jiang2023structgpt}
Jiang, J.; Zhou, K.; Dong, Z.; Ye, K.; Zhao, W.~X.; and Wen, J.-R. 2023.
\newblock StructGPT: A General Framework for Large Language Model to Reason
  over Structured Data.
\newblock In \emph{Proceedings of the 2023 Conference on Empirical Methods in
  Natural Language Processing}, 9237--9251.

\bibitem[{Kong et~al.(2024)Kong, Zhang, Shen, Srinivasan, Lei, Faloutsos,
  Rangwala, and Karypis}]{kong2024opentab}
Kong, K.; Zhang, J.; Shen, Z.; Srinivasan, B.; Lei, C.; Faloutsos, C.;
  Rangwala, H.; and Karypis, G. 2024.
\newblock OpenTab: Advancing Large Language Models as Open-domain Table
  Reasoners.
\newblock \emph{CoRR}, abs/2402.14361.

\bibitem[{Kweon et~al.(2023)Kweon, Kwon, Cho, Jo, and Choi}]{kweon2023open}
Kweon, S.; Kwon, Y.; Cho, S.; Jo, Y.; and Choi, E. 2023.
\newblock Open-WikiTable: Dataset for Open Domain Question Answering with
  Complex Reasoning over Table.
\newblock In \emph{Findings of the Association for Computational Linguistics:
  ACL 2023}, 8285--8297.

\bibitem[{Pal et~al.(2023)Pal, Yates, Kanoulas, and
  de~Rijke}]{pal2023multitabqa}
Pal, V.; Yates, A.; Kanoulas, E.; and de~Rijke, M. 2023.
\newblock MultiTabQA: Generating Tabular Answers for Multi-Table Question
  Answering.
\newblock In \emph{Proceedings of the 61st Annual Meeting of the Association
  for Computational Linguistics (Volume 1: Long Papers)}, 6322--6334.

\bibitem[{Pasupat and Liang(2015)}]{pasupat2015compositional}
Pasupat, P.; and Liang, P. 2015.
\newblock Compositional Semantic Parsing on Semi-Structured Tables.
\newblock In \emph{Proceedings of the 53rd Annual Meeting of the Association
  for Computational Linguistics and the 7th International Joint Conference on
  Natural Language Processing (Volume 1: Long Papers)}, 1470--1480.

\bibitem[{Qin et~al.(2024)Qin, Wei, Chen, Zhou, Huang, Si, Lu, and
  Che}]{qin2024croprompt}
Qin, L.; Wei, F.; Chen, Q.; Zhou, J.; Huang, S.; Si, J.; Lu, W.; and Che, W.
  2024.
\newblock CroPrompt: Cross-task Interactive Prompting for Zero-shot Spoken
  Language Understanding.
\newblock \emph{arXiv preprint arXiv:2406.10505}.

\bibitem[{Qin et~al.(2022)Qin, Wei, Ni, Zhang, Che, Li, and Liu}]{qin2022multi}
Qin, L.; Wei, F.; Ni, M.; Zhang, Y.; Che, W.; Li, Y.; and Liu, T. 2022.
\newblock Multi-domain spoken language understanding using domain-and
  task-aware parameterization.
\newblock \emph{Transactions on Asian and Low-Resource Language Information
  Processing}, 21(4): 1--17.

\bibitem[{Qin et~al.(2021)Qin, Xie, Che, and Liu}]{qin2021survey}
Qin, L.; Xie, T.; Che, W.; and Liu, T. 2021.
\newblock A Survey on Spoken Language Understanding: Recent Advances and New
  Frontiers.
\newblock In \emph{Proceedings of the Thirtieth International Joint Conference
  on Artificial Intelligence, {IJCAI} 2021, Virtual Event / Montreal, Canada,
  19-27 August 2021}, 4577--4584. ijcai.org.

\bibitem[{Ramshaw and Marcus(1999)}]{ramshaw1999text}
Ramshaw, L.~A.; and Marcus, M.~P. 1999.
\newblock Text chunking using transformation-based learning.
\newblock In \emph{Natural language processing using very large corpora},
  157--176. Springer.

\bibitem[{Robertson and Zaragoza(2009)}]{robertson2009probabilistic}
Robertson, S.; and Zaragoza, H. 2009.
\newblock The Probabilistic Relevance Framework: BM25 and Beyond.
\newblock \emph{Foundations and Trends{\textregistered} in Information
  Retrieval}, 3(4): 333--389.

\bibitem[{Robertson and Walker(1994)}]{robertson1994some}
Robertson, S.~E.; and Walker, S. 1994.
\newblock Some simple effective approximations to the 2-poisson model for
  probabilistic weighted retrieval.
\newblock In \emph{SIGIR’94: Proceedings of the Seventeenth Annual
  International ACM-SIGIR Conference on Research and Development in Information
  Retrieval, organised by Dublin City University}, 232--241. Springer.

\bibitem[{Shwartz-Ziv and Armon(2022)}]{shwartz2022tabular}
Shwartz-Ziv, R.; and Armon, A. 2022.
\newblock Tabular data: Deep learning is not all you need.
\newblock \emph{Information Fusion}, 81: 84--90.

\bibitem[{Sui et~al.(2023)Sui, Zou, Zhou, He, Du, Han, and
  Zhang}]{sui2023tap4llm}
Sui, Y.; Zou, J.; Zhou, M.; He, X.; Du, L.; Han, S.; and Zhang, D. 2023.
\newblock Tap4llm: Table provider on sampling, augmenting, and packing
  semi-structured data for large language model reasoning.
\newblock \emph{arXiv preprint arXiv:2312.09039}.

\bibitem[{Yang et~al.(2024)Yang, Yang, Hui, Zheng, Yu, Zhou, Li, Li, Liu, Huang
  et~al.}]{yang2024qwen2}
Yang, A.; Yang, B.; Hui, B.; Zheng, B.; Yu, B.; Zhou, C.; Li, C.; Li, C.; Liu,
  D.; Huang, F.; et~al. 2024.
\newblock Qwen2 technical report.
\newblock \emph{arXiv preprint arXiv:2407.10671}.

\bibitem[{Yu et~al.(2018)Yu, Zhang, Yang, Yasunaga, Wang, Li, Ma, Li, Yao,
  Roman et~al.}]{yu2018spider}
Yu, T.; Zhang, R.; Yang, K.; Yasunaga, M.; Wang, D.; Li, Z.; Ma, J.; Li, I.;
  Yao, Q.; Roman, S.; et~al. 2018.
\newblock Spider: A Large-Scale Human-Labeled Dataset for Complex and
  Cross-Domain Semantic Parsing and Text-to-SQL Task.
\newblock In \emph{Proceedings of the 2018 Conference on Empirical Methods in
  Natural Language Processing}, 3911--3921.

\bibitem[{Zhong, Xiong, and Socher(2017)}]{zhong2017seq2sql}
Zhong, V.; Xiong, C.; and Socher, R. 2017.
\newblock Seq2sql: Generating structured queries from natural language using
  reinforcement learning.
\newblock \emph{arXiv preprint arXiv:1709.00103}.

\bibitem[{Zhu et~al.(2024)Zhu, Cheng, An, Wang, Chen, and Huang}]{zhu2024zero}
Zhu, Z.; Cheng, X.; An, H.; Wang, Z.; Chen, D.; and Huang, Z. 2024.
\newblock Zero-Shot Spoken Language Understanding via Large Language Models: A
  Preliminary Study.
\newblock In \emph{Proceedings of the 2024 Joint International Conference on
  Computational Linguistics, Language Resources and Evaluation (LREC-COLING
  2024)}, 17877--17883.

\end{thebibliography}

\newpage

\section{Appendix}

\begin{figure*}[htbp]
 \centering
 \includegraphics[width=1.0\textwidth]{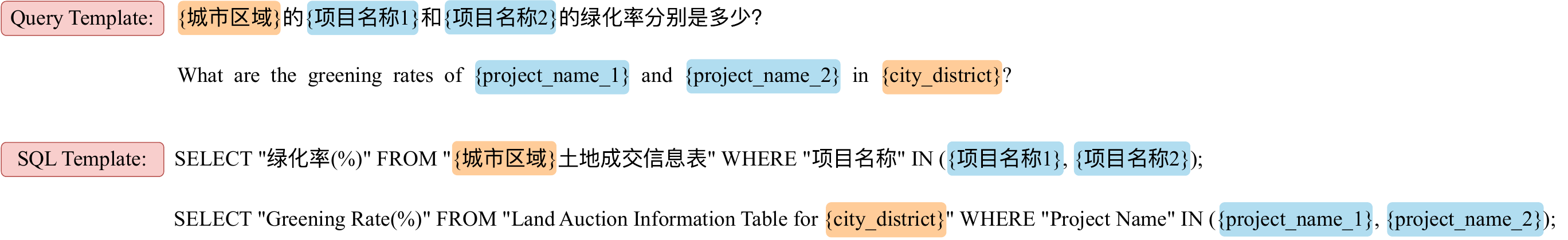}
 \centering
 \caption{Showcase of Template Filling.}
 \label{fig:Template_Filling}
\end{figure*}

\subsection{A. Template Filling}

\begin{figure}[htbp]
 \centering
 \includegraphics[width=0.9\columnwidth]{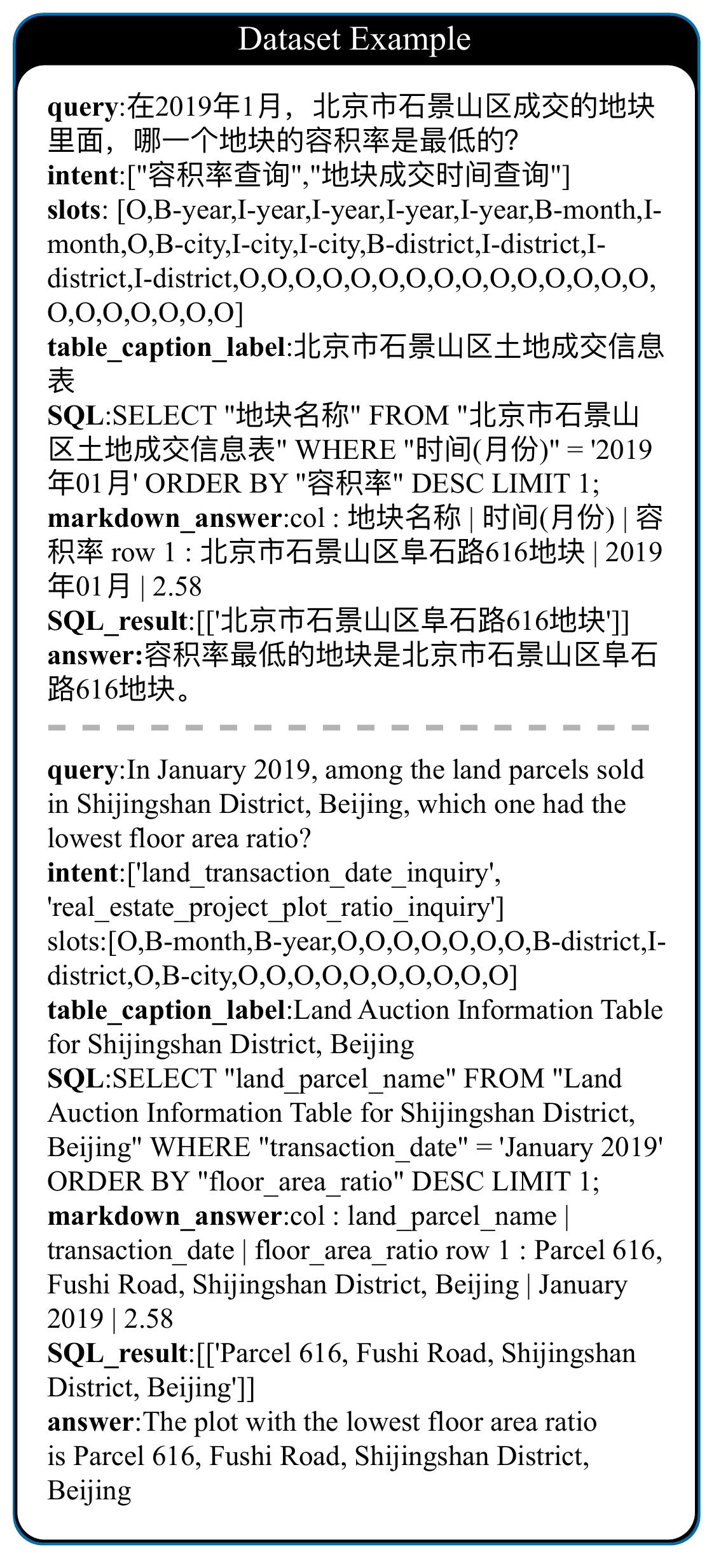}
 \centering
 \caption{Showcase of a Example in the Dataset.}
 \label{fig:QA_example}
\end{figure}

\begin{figure}[htbp]
 \centering
 \includegraphics[width=3.0in]{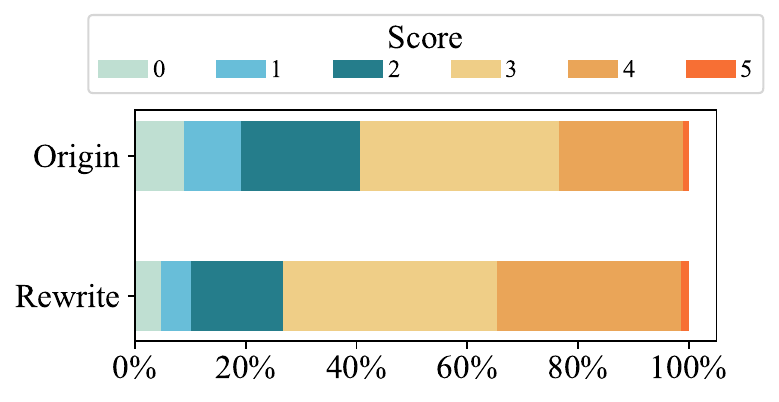}
 \centering
 \caption{Visualizing Distribution of Rewrite Scores}
 \label{fig:Rewrite Scores}
\end{figure}

We define 16 types of intents based on the column headers from collected real estate tables in three major domains: property information, real estate company finance, and land auction information. Around these intents, we create 90 templates for generating a dataset. These templates cater to two scenarios: single-table and multi-table queries. For single-table queries, where the answer is confined to a single table, we develop 67 templates. For multi-table queries, where the answer may span multiple tables, we create 23 templates. These templates are then populated to generate question-answer pairs.

QA pairs are automatically generated using the 90 templates mentioned. The complete details of these templates will be published on GitHub after the anonymity period. Below, we present an example template to illustrate our approach. As shown in Figure \ref{fig:Template_Filling}, each template pair consists of two components: a query template and an SQL template, both requiring specific variables to be filled. In the query template, shaded areas indicate positions to be filled with variables. For instance, the blue-shaded $\{project\_name\_1\}$ and $\{project\_name\_2\}$ represent project names, while the yellow-shaded $\{city\_district\}$ indicates a city district. The unshaded tokens are fixed components of the query template. The SQL template follows a similar structure, with corresponding shaded areas for variables.

All variables are specifically selected from the collected real estate tables. The example in Figure \ref{fig:Template_Filling} pertains to land auction information. We first extract all city districts from relevant table captions to create a set of city districts. We then randomly select one city district and insert it into the $\{city\_district\}$ slots in both the query and SQL templates. After establishing the city district, the corresponding table caption is determined. We then randomly select two project names from the target table and insert them into the $\{project\_name\_1\}$ and $\{project\_name\_2\}$ positions in both templates, generating the final question and its SQL query.

The intent is determined by the fixed phrases in the template, while the slot information is labeled using the variable names. Finally, we input the SQL query execution results along with the original query into a LLM to generate a natural language answer. This process completes the population of a single template pair. After extensive sampling, we filter out valid QA pairs with non-empty SQL execution results for further use.

\begin{table}[htp]\small
\begin{tabular}{ll}
\hline
Statistics                                         & Num    \\ \hline
\# Utterances                                       & 20762  \\
\# Single Table                                     & 15002  \\
\# Multi Tables                                     & 5760   \\
\# Single Intent                                    & 17476  \\
\# Multi Intents                                    & 3286   \\
\# Ave, Words per Utterance                         & 40     \\
\# Intents                                          & 16     \\
\# Slots                                            & 6      \\
\# Ave, Slots                                       & 3.06   \\
\# Ave, Rows                                        & 252.87 \\
\# Ave, Columns                                     & 7.78   \\
\# Poperty Information Tables                       & 4825   \\
\# Real Estate Company Finance Information   Tables & 4      \\
\# Land Auction Information Tables                  & 103    \\
\# Factual Questions                                & 12340  \\
\# Inferential Questions                            & 1015   \\
\# Comparative Questions                            & 7407   \\ \hline
\end{tabular}
\caption{Dataset Statistics}\label{Dataset statics}
\end{table}

\begin{table}[htp]\small
\begin{tabular}{ccccccc}
\hline
\multirow{2}{*}{Model} & \multicolumn{3}{c}{Intent}                       & \multicolumn{3}{c}{Slots}                        \\ \cline{2-7} 
                       & P              & R              & F1             & P              & R              & F1             \\ \hline
Qwen2 7b               & 91.99          & 93.78          & 92.87          & 97.14          & 98.29          & 97.71          \\
Qwen2 72b              & 97.97          & 98.89          & 98.43          & 97.14          & 99.26          & 98.19          \\
GLM4 9b                & 93.36          & 95.75          & 94.54          & 95.29          & 97.86          & 96.56          \\
BERT                   & \textbf{99.04} & \textbf{99.62} & \textbf{99.33} & \textbf{99.54} & \textbf{99.31} & \textbf{99.42} \\ \hline
\end{tabular}
\caption{Intent and Slot Prediction Results}\label{Compare result of SLU predict}
\end{table}

A complete example of RETQA is illustrated in Figure \ref{fig:QA_example}. For each query, we provide both the intent label(s) and the slot labels, utilizing the BIO tagging format for the slots. Additionally, standardized table captions are provided to facilitate the execution of retrieval tasks. All standard SQL statements have been verified for correct execution and are accompanied by their corresponding execution results (SQL\_result). The markdown\_answer can be used for end-to-end Markdown input-output tasks.

\subsection{B. Rewrite Result}
Figure \ref{fig:Rewrite Scores} presents the scoring results for queries before and after rewriting, using DeepSeek-V1 (Dense-67B). The scoring criteria evaluate the extent to which each query resembles natural language as opposed to template-generated text, with scores ranging from 0 to 5. A score of 5 indicates a strong resemblance to human language patterns, while a score of 0 suggests a more template-like generation. Different colors represent the percentage distribution of each score across the entire dataset, displayed from 0 to 5, left to right. The distribution shows that after rewriting by LLMs, the scores shift to the right, indicating that the rewritten queries more closely align with natural human language.

\subsection{C. Statistics}

Table \ref{Intents distribution} presents the statistical count of queries under 16 intent classifications. For queries containing multiple intents, the intent labels are connected using a plus sign. Table \ref{Dataset statics} provides a statistical overview of various aspects of the dataset, where ``Ave'' is an abbreviation for ``Average''. We have enumerated the total number of QA pairs in the dataset, as well as detailed information such as the number of QA pairs belonging to single-intent and multi-intent categories. It is specifically noted that ``\# Ave, Rows'' represents the average number of rows in the target table corresponding to each question, excluding the header row. When a question involves multiple tables, the row counts are cumulatively calculated.

\subsection{D. SLU Predict}

Table \ref{Compare result of SLU predict} presents the performance comparison of Intent and Slot prediction in queries using BERT and ICL methods. For the BERT model, we fine-tune the ``Bert-base-chinese\footnote[1]{https://huggingface.co/google-bert/bert-base-chinese}'' version on the training set, following the procedure outlined in the main text. In the ICL scenario, we select a sufficient number of samples from the training set, covering all intent types, to construct the context. These samples included both intent and slot information, enabling LLMs to generate intent and slot labels for new queries in a single pass. When the SLU label format in the LLM output is not standardized, we select the label that most closely matches the output characters. The experimental results indicate that the fine-tuned BERT model achieves better performance in predicting SLU labels compared with ICL methods. However, fine-tuning requires extensive manual annotation, whereas the ICL methods can achieve comparable results with only a few examples.

\begin{table*}[htbp]
\centering
\begin{tabular}{lc}
\hline
Intent                                                                                         & Num   \\ \hline
real\_estate\_project\_sales\_volume\_query                                                             & 1465  \\
real\_estate\_project\_developer\_information\_query                                                    & 207   \\
real\_estate\_project\_average\_price\_query                                                            & 2367  \\
real\_estate\_project\_green\_coverage\_ratio\_query                                                    & 1076  \\
real\_estate\_project\_building\_density\_query                                                         & 1135  \\
real\_estate\_project\_plot\_ratio\_query                                                               & 1058  \\
land\_total\_price\_query                                                                               & 1356  \\
land\_ownership\_query                                                                                  & 448   \\
land\_transaction\_date\_query                                                                          & 501   \\
company\_cost\_query                                                                                    & 1894  \\
company\_nature\_query                                                                                  & 516   \\
company\_risk\_assessment\_query                                                                        & 732   \\
company\_debt\_query                                                                                    & 1274  \\
company\_total\_revenue\_query                                                                          & 1534  \\
company\_profit\_query                                                                                  & 1338  \\
company\_debt\_default\_query                                                                           & 575   \\
real\_estate\_project\_sales\_volume\_query+   real\_estate\_project\_sales\_volume\_query              & 1794  \\
land\_transaction\_date\_query +   real\_estate\_project\_plot\_ratio\_query                            & 591   \\
land\_transaction\_date\_query +   land\_total\_price\_query                                            & 300   \\
real\_estate\_project\_green\_coverage\_ratio\_query   +real\_estate\_project\_building\_density\_query & 287   \\
land\_total\_price\_query +   real\_estate\_project\_plot\_ratio\_query                                 & 270   \\
company\_debt\_query +   company\_risk\_assessment\_query                                               & 28    \\
company\_debt\_default\_query +   company\_debt\_query                                                  & 16    \\ \hline
total                                                                                                   & 20762 \\ \hline
\end{tabular}
\caption{Intents distribution statistics.}
\label{Intents distribution}
\end{table*}

\subsection{E. Computing Infrastructure Statement}

During the training of Bert-base-chinese for SLU prediction, we use a single GeForce RTX 4090 GPU. All neural networks are implemented using PyTorch\footnote[2]{https://pytorch.org/}, with version 2.3.1. For in-context learning, we utilize six NVIDIA A800-SXM4-80GB GPUs and employ the vLLM library\footnote[3]{https://docs.vllm.ai/} to execute LLM inference tasks. Of these GPUs, four are dedicated to running the Qwen2 72B model, one to the GLM4 9B model, and another to the Qwen2 7B model.

\subsection{F. Discussion}

Since LLMs rely on context information for ICL, we have included all prompt information in the supplementary materials under the SLUTQA directory as .json files. Due to the extended runtime required for QA using LLMs, we were unable to complete multiple rounds of experiments before the deadline to fully support our research findings as specified in the checklist, including the number of algorithm runs, mean and variance statistics, and significance tests. However, this does not diminish the validity of our final conclusions. We plan to supplement these detailed experimental data in the final version.

\end{document}